\documentclass[a4paper]{article}%
\usepackage{placeins}
\usepackage{algorithm}
\usepackage{algorithmic}
\usepackage{amsthm}
\usepackage{amssymb}
\usepackage{graphicx}
\usepackage{rotating}
\usepackage{amsfonts}
\usepackage{mathrsfs}
\usepackage{amsmath}
\usepackage{babel}
\usepackage{lastpage}
\usepackage[utf8]{inputenc}
\usepackage[legalpaper,bookmarks=true,colorlinks=true,linkcolor=blue,citecolor=blue]{hyperref}
\usepackage{graphicx}%
\usepackage{fancyhdr,enumerate}
\usepackage{color}

\usepackage[mathlines]{lineno}
\usepackage{lscape}
\usepackage{epsfig}
\usepackage[sort&compress]{natbib}
\usepackage{epstopdf}
\usepackage{subfigure}
\usepackage{float,adjustbox}
\usepackage{bm}
\usepackage{booktabs}
\usepackage{geometry}
\usepackage{geometry}
\begin{document}

\title{Enhancing Credit Default Prediction Using  Boruta Feature Selection and DBSCAN Algorithm with Different Resampling Techniques}

\author{Obu-Amoah Ampomah$^{\textbf{1}}$, Edmund Agyemang$^{\textbf{2,3*}}$,  Kofi Acheampong$^{\textbf{4}}$, Louis Agyekum$^{\textbf{5}}$\\
$^{\textbf{1}}$Department of Statistics, Western Michigan University, Kalamazoo, Michigan.\\ 
$^{\textbf{2}}$Department of Biostatistics and Data Science, Celia Scott Weatherhead School of \\Public Health and Tropical Medicine, Tulane University, New Orleans, Louisiana, USA.\\
$^{\textbf{3}}$School of Mathematical and Statistical Science, College of Sciences,\\ University of Texas Rio Grande Valley, Edinburg, Texas, USA.\\
$^{\textbf{4}}$Department of Economics, Western Michigan University, Kalamazoo, Michigan.\\
$^\textbf{{5}}$Department of Economics, University of Ottawa, Ottawa, ON K1N 6N5, Canada.\\
Corresponding author: \color{blue}edmundfosu6@gmail.com, ORCID ID: 0000-0001-8124-4493}
		\date{}
		\maketitle

\begin{abstract}
\noindent	
This study examines credit default prediction by comparing three techniques namely SMOTE, SMOTE-Tomek, and ADASYN that are commonly used to address the class imbalance problem in  credit default situations. Recognizing that credit default datasets are typically skewed, with defaulters comprising a much smaller proportion than non-defaulters, we began our analysis by evaluating machine learning (ML) models on the imbalanced data without any resampling to establish baseline performance. These baseline results provide a reference point for understanding the impact of subsequent balancing methods. In addition to traditional classifiers such as Naive Bayes and K-Nearest Neighbors (KNN), our study also explores the suitability of advanced ensemble boosting algorithms including Extreme Gradient Boosting (XGBoost), AdaBoost, Gradient Boosting Machines (GBM), and LightGBM for credit default prediction using Boruta feature selection and DBSCAN-based outlier detection, both before and after resampling. A real-world credit default data set sourced from the University of Cleveland ML Repository was used to build ML classifiers, and their performances were tested. The  criteria chosen to measure model performance are the area under the receiver operating characteristic curve (ROC-AUC), area under the precision-recall curve (PR-AUC), G-mean and F1-scores. The results from this empirical study indicate that the Boruta+DBSCAN+SMOTE-Tomek+GBM classifier outperformed the other ML models (F1-score: 82.56\%, G-mean: 82.98\%, ROC-AUC: 90.90\%, PR-AUC: 91.85\%) in a credit default context following an initial training using the 5-fold cross-validation with F1-score as a scoring criterion coupled with extensive hyperparameter tuning. The efficiency of ensemble boosting techniques as a useful weapon in the fight against credit default has been highlighted in this study. The findings establish a foundation for future progress in creating more resilient and adaptive credit default systems, which will be essential as credit-based transactions continue to rise worldwide.\\

\end{abstract}

\section{Introduction}	
As credit-based transactions continue to rise globally, the prediction of credit card default has become a central concern for financial institutions \cite{farabi2024enhancing}.
The ability to accurately anticipate whether a borrower will meet their repayment obligations is critical not just for protecting lenders from financial losses, but also for ensuring the broader stability of financial systems \cite{brown2012experimental}. Defaults can have cascading effects, straining institutional resources and potentially contributing to wider economic downturns. As such, reliable default prediction systems are essential in credit risk management, loan approval processes, and regulatory compliance. Despite advances in data analytics and machine learning, default prediction remains a complex and challenging task. Credit default datasets are often high-dimensional, with numerous features ranging from demographics, balance limits to transaction histories. This abundance of features can introduce noise and redundancy, making it difficult to identify truly informative predictors \cite{cunha2025noise}. Additionally, there is often a problem with class imbalance: events of default happen much less often than non-defaults, which makes it harder for standard classifiers to work well and can lead to unfair predictions when solely relying on metrics like accuracy. \\

\noindent
Issues concerning data quality ranging from missing values, outliers, and inconsistent records further hamper model accuracy \cite{kiebler2022preliminary}. In addition, balancing predictive power with model interpretability and explainable artificial intelligence is an ongoing challenge, as stakeholders require transparent models to ensure fair and accountable decision-making \cite{nortey2025ai}. As a matter of fact, in the wake of addressing these challenges, researchers and practitioners are exploring advanced techniques such as best preprocessing techniques,  ensemble, boosting, hybrid methods and robust anomaly detection, which can help improve model performance and robustness \cite{lessmann2015benchmarking,khalid2024enhancing}. It is in this regard that this current study comes in handy to aid in the refinement of credit default through the application of Boruta; an extensive feature selection approach and Density-Based Spatial Clustering of Applications with Noise (DBSCAN); a rigorous outlier detection algorithm coupled with data balancing techniques like Synthetic Minority Over-sampling Technique (SMOTE), SMOTE combined with Tomek links (SMOTE-Tomek) and Adaptive Synthetic Sampling (ADASYN). To further boost the predictive accuracy of the Boruta+DBSCAN balanced data, ensemble boosting algorithms such as Gradient Boosting Machines (GBM), Light Gradient Boosting Machines (Light GBM), Extreme Gradient Boosting (XGBoost), Adaptive Boosting in addition to Naive Bayes (NB) and K-Nearest Neighbors (KNN) are applied. \\

\noindent
In financial datasets such as credit defaults, where variables may be abundant and frequently redundant, feature selection aims to concentrate the study on the most informative attributes, hence enhancing decision-making. Among various feature selection techniques, the Boruta algorithm has emerged as a powerful tool for identifying all relevant features. Boruta is a wrapper-based feature selection method that builds on the random forest classifier \cite{kursa2010feature}. Compared to other methods such as recursive feature elimination (RFE) or filter-based approaches, Boruta offers advantages in robustness and the ability to handle interactions among features \cite{degenhardt2019evaluation}. It is particularly well-suited for datasets with complex, nonlinear relationships and a large number of features. In the context of credit default prediction, Boruta has been successfully used to streamline feature sets, leading to more efficient model training and improved generalizability \cite{hegde2023feature, ganiyu2024credit}. As the financial industry increasingly relies on data-driven insights, the use of advanced feature selection methods like Boruta algorithm becomes essential for constructing reliable and interpretable machine learning systems. Hence, its adoption in this study to aid in the refinement of credit default prediction.\\

\noindent
Credit default datasets are most often characterized by complex, nonlinear patterns and may contain significant numbers of outliers due to unusual customer behaviors, errors, or fraud. Effectively identifying and handling these anomalies can improve the performance and reliability of subsequent predictive models \cite{aggarwal2017introduction}. In lieu of this, the DBSCAN, a well known clustering algorithm was employed as a tool in this study for noise reduction and outlier detection. In contrast with traditional clustering techniques like k-means, which require prior determination of cluster numbers and exhibit sensitivity to cluster morphology, DBSCAN automatically creates clusters of irregular shapes and efficiently detects noise points or outliers \cite{bushra2024autoscan}. This makes it especially valuable for preprocessing tasks such as filtering out extreme or unusual default or  non-default cases before applying supervised learning models \cite{Schubert2017dbscan}. In the field of default prediction, employing DBSCAN for anomaly identification serves two primary purposes. DBSCAN isolates atypical transactions or profiles that could adversely affect model training, and it also offers insights into rare yet significant behaviors that may signal possible defaults.\\

\noindent
Recent studies have demonstrated that combining DBSCAN with supervised ML pipelines can enhance both model robustness and interpretability, ultimately leading to more reliable predictions in risk-sensitive applications. For example, \cite{machado2022assessing} illustrated that employing a hybrid ML methodology, wherein commercial customers are initially clustered using techniques such as DBSCAN and diverse predictive models are subsequently applied to each cluster, significantly enhances the precision of credit risk prediction relative to conventional methods alone.  DBSCAN shown efficacy in discerning natural clusters within customer data, with optimal outcomes achieved through its integration with tree-based models like Random Forest or Decision Tree, particularly when historical credit score data was used.  The results illustrate that hybrid approach, especially with DBSCAN, provides an effective novel method for evaluating credit risk in commercial banking. \cite{machado2022applying} also adopted a hybrid ML approach to predict customer risk-adjusted revenue (RAR) in the P2P lending industry, using a large dataset from Lending Club. Individual ML models (like Gradient Boosting, Random Forest, Decision Tree, SVM, ANN, and Adaboost)  and hybrid frameworks that first cluster customers using unsupervised algorithms specifically k-Means and DBSCAN before applying supervised regressors were employed. DBSCAN notably offers strong validation metrics (e.g., $R^2$ above 85\% for many regressors) and offers an additional advantage by identifying irregular groupings and noise in consumer data. The study reveals generally that DBSCAN combined with tree-based models provides strong prediction for RAR and that using clustering prior to regression enables better segmentation and more accurate, efficient customer value assessment in financial services. \\

\noindent
Owing to the promising results of Boruta, DBSCAN and oversampling techniques (SMOTE, SMOTE-Tomek and ADASYN), this study aims to evaluate the performance of several ML models including Naive Bayes, KNN, XGBoost, AdaBoost, GBM and Light GBM in predicting credit card default for financial service applications. The remainder of the paper is organized as follows: Section \ref{Sec2} discusses the data and methods used for the study. Section \ref{Sec3} discusses the results and findings of the study whilst section \ref{Sec4} concludes the study and provide recommendations for further work.

\section{Data and Methods \label{Sec2}}
The data and methods adopted for the study are discussed in this section.

\subsection{Data Description}
The data is available on University of Cleveland Machine Learning Repository provided by \cite{default_of_credit_card_clients_350} and can be freely assessed at \url{https://archive.ics.uci.edu/dataset/350/default+of+credit+card+clients}. In this study, the response variable is a binary indicator of credit default, where a value of $1$ denotes a default (``Yes'') and $0$ indicates no default (``No''). This data considers the following $23$ variables as explanatory factors:

\begin{itemize}
	\item[\textbf{
		X1}:] Also \textbf{LIMIT\_BAL} is the amount of the given credit (in NT dollars), which encompasses both the individual's consumer credit and any associated family (supplementary) credit.
	\item[\textbf{X2}:] Sex (1 = male; 2 = female).
	\item[\textbf{X3}:] Education level (1 = graduate school; 2 = university; 3 = high school; 4 = others).
	\item[\textbf{X4}:] Marital status (1 = married; 2 = single; 3 = others).
	\item[\textbf{X5}:] Age (in years).
	\item[\textbf{X6--X11}:] Also \textbf{PAY\_0 - PAY\_6} is the history of past payment. These variables represent the repayment status for each month from April to September 2005 as follows: \textbf{X6} = September 2005, \textbf{X7} = August 2005, $\ldots$, \textbf{X11} = April 2005. Repayment status is coded as: $-1$ (paid duly); $1$ (payment delay for one month); $2$ (delay for two months); $\ldots$; $8$ (delay for eight months); $9$ (delay for nine months or more).
	\item[\textbf{X12--X17}:] Also  \textbf{BILL\_AMT1 - BILL\_AMT6
		 } are the amount of bill statements (in NT dollars) for each month: \textbf{X12} = September 2005, \textbf{X13} = August 2005, $\ldots$, \textbf{X17} = April 2005.
	\item[\textbf{X18--X23}:] Also \textbf{PAY\_AMT1 - PAY\_AMT6} are the amount of previous payments (in NT dollars) for each month: \textbf{X18} = payment made in September 2005, \textbf{X19} = August 2005, $\ldots$, \textbf{X23} = April 2005.
\end{itemize}

\begin{figure}[!hbtp]
	\centering
	\includegraphics[width=1.0\linewidth]{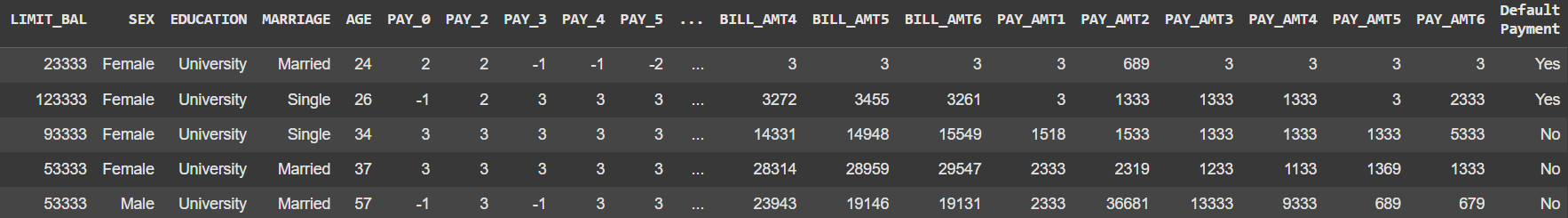}
	\caption{\textbf{Extract of the Credit Card Default data}}
	\label{Data}
\end{figure}

\noindent
Figure \ref{Data} presents an extract of the data used for the study.

\subsection{Data Preprocessing}
First, the credit default data was examined for missing values using Pandas in Python. No missing values were recorded given the type of the data, hence imputation techniques were not needed. This was verified by means of \textit{dataset.isnull()} method, which yielded zero for every feature, so indicating the absence of missing data.
In this study, the Boruta algorithm \cite{kursa2010feature} was employed to identify the most relevant predictors of credit default. First, the dataset was imported into a \texttt{pandas} DataFrame and split into the feature matrix \(X\) and the binary target vector \(y\) (“Default”). All categorical variables in \(X\) were one‐hot encoded resulting in a fully numeric matrix \(X_{\mathrm{enc}}\). A \texttt{RandomForestClassifier} was then instantiated with balanced class weights, a maximum tree depth of 5, parallel computation, and a fixed random seed for reproducibility. This classifier served as the underlying estimator for \texttt{BorutaPy}, which we configured to determine the number of trees automatically (\texttt{n\_estimators=`auto'}) and to report progress verbosely.\\

\newpage
\noindent
During fitting, Boruta iteratively augments the data with shadow features (permuted copies of the originals), assesses feature importance via the random forest, and confirms or rejects each original feature based on comparisons with its best-performing shadow counterpart. Finally, the confirmed features using the selector’s \texttt{support\_} attribute and constructed a reduced dataset \(X_{\mathrm{sel}}\) containing only these variables were extracted. After successful implementation, the Boruta algorithm successfully selected 20 features (with the exception of sex, education and marital status which were removed after complete execution) out of the initial 23 features. These 20 selected Boruta features were then subjected to  the DBSCAN algorithm to aid in the removal of any outliers. After successfully removing outliers via DBSCAN, the Boruta+DBSCAN outlier free data was then balanced by the three (3) oversampling techniques considered in the study: SMOTE, SMOTE-Tomek and ADASYN (see \cite{agbota2024enhancing,agyemang2025addressing} for details on SMOTE, SMOTE-Tomek and ADASYN oversampling techniques). The final balanced data was then analyzed using the six (6) machine learning (ML) algorithms discussed in the subsequent sections. However, before employing the credit default classifiers, the data was segmented into training (80\%) and testing (20\%) subsets. The train data was utilized to develop the six (6)  ML models, while the test data served to assess these models' predictive accuracy. Before training, z-score normalization was utilize in scaling to make all features contribute equally to the result.

\subsection{Naive Bayes Classifier}
Naive Bayes classifiers are effective instruments for detecting credit  default and non-default, owing to their simplicity and capability to manage extensive datasets with multiple features. This method uses Bayes theorem to determine the likelihood of a customer defaulting based on its features, including transaction amount, limit balance and location. Despite its “naive" assumption of feature independence, which may not be applicable in practical situations, Naive Bayes classifiers can nonetheless yield credible outcomes by separately calculating the probabilities of each feature. Naive Bayes classifier facilitates swift identification of anomalies in credit default prediction. This effectiveness is particularly important in today's fast-paced financial environment, where timely detection of credit default activity can save institutions significant amounts of money and protect consumers. As a result, the integration of Naive Bayes classifiers into credit default prediction continues to gain traction among financial organizations \cite{ismail2023review}.

\color{black}
\subsection{K-Nearest Neighbors (KNN)} 
In this study, the Manhattan distance (which emerged as the best metric hyperparameter) was used in the KNN algorithm to measure the closeness between data points in the credit  default data. Unlike the Euclidean metric, Manhattan distance calculates the sum of absolute differences across dimensions, as shown in (\ref{d1}).

\begin{equation}
	d(a, b) = \sum_{i=1}^{n} |a_i - b_i|
	\label{d1}
\end{equation}
Here, $a$ and $b$ are points in an $n$-dimensional feature space. The parameter $k$ denotes the number of nearest neighbors considered for classification. Optimal values of $k$ were determined through cross-validation.

\subsubsection{Classification Decision Rule}
To classify a new point $x$, the process includes:

\begin{enumerate}
	\item [(i)] Compute Manhattan distances between $x$ and all training points.
	\item [(ii)] Select the $k$ nearest neighbors.
	\item [(iii)] Assign the class label by majority vote, as given in (\ref{d11}):
\end{enumerate}

\begin{equation}
	\hat{y} = \arg \max_{c \in C} \sum_{i=1}^{k} \mathbf{1}(y_i = c)
	\label{d11}
\end{equation}
This approach is simple, effective, and well-suited for the classification of credit default behavior.

\subsection{Extreme Gradient Boosting (XGBoost) Classification}
XGBoost is a high-performance implementation of gradient boosting tailored for both speed and accuracy. For the classification of credit  default behavior, it builds an ensemble of decision trees, where each tree corrects errors from the previous ones \cite{nortey2025ai}. The XGBoost model is defined in (\ref{XG1}) as:

\begin{equation}
F_{m}(x) = \sum_{k=1}^{m} h_k(x)
\label{XG1}
\end{equation}
Here, \( h_k(x) \) denotes the prediction from the \(k\)-th tree, and \( F_m(x) \) is the final output after \( m \) iterations.\\

\noindent
The learning process minimizes the regularized objective function given in (\ref{h4}) by:

\begin{equation}
	\mathcal{L}(\theta) = \sum_{i=1}^{n} L(y_i, \hat{y}_i) + \sum_{k=1}^{m} \Omega(h_k)
	\label{h4}
\end{equation}
where \( L(y_i, \hat{y}_i) \) is the classification loss (e.g., logistic loss), and \( \Omega(h_k) = \gamma T + \frac{1}{2} \lambda \sum_{j=1}^{T} w_j^2 \) is a regularization term to control model complexity. Here, \( T \) is the number of leaves in a tree, \( w_j \) are the leaf weights, and \( \gamma, \lambda \) are regularization parameters. XGBoost effectively handles class imbalance and overfitting, making it suitable for the classification of credit default.

\subsection{Adaptive Boosting (AdaBoost) Classification}
AdaBoost is an ensemble method that improves classification performance by combining multiple weak learners. In each iteration, it focuses more on misclassified instances by increasing their weights \cite{nortey2025ai}. The final Adaboost model is given in (\ref{h5}) by:

\begin{equation}
	F_{m}(x) = \sum_{k=1}^{m} \alpha_k h_k(x)
	\label{h5}
\end{equation}
Here, \( h_k(x) \) is the prediction from the \( k \)-th weak learner, and \( \alpha_k \) is the weight assigned based on its classification accuracy. The learner weight \(\alpha_k \) is calculated using the error rate \( e_k \) given in (\ref{h6}) as:

\begin{equation}
	\alpha_k = \frac{1}{2} \ln\left(\frac{1 - e_k}{e_k}\right)
	\label{h6}
\end{equation}
This adaptive process allows AdaBoost to sequentially improve classification accuracy, making it suitable for predicting credit default behavior with increased sensitivity to difficult cases.

\subsection{Gradient Boosting Machines (GBM) Classification}
Gradient Boosting Machines (GBM) build an ensemble of decision trees sequentially, where each tree is trained to correct the errors of the previous one. For classification tasks such as predicting default status, GBM optimizes a differentiable loss function through gradient descent.
The additive model is defined in (\ref{h7}) as:

\begin{equation}
	F_{m}(x) = \sum_{k=1}^{m} h_k(x)
	\label{h7}
\end{equation}
where \( h_k(x) \) represents the \( k \)-th decision tree, and \( F_m(x) \) is the cumulative prediction after \( m \) iterations. Each tree is trained on the negative gradient of the loss function with respect to the previous prediction, typically using the logistic loss for binary classification. This approach enables GBM to improve performance iteratively and handle non-linear relationships in the data. GBM is effective in modeling complex patterns in the default data and provides high predictive accuracy when tuned properly.

\subsection{Light Gradient Boosting Machine (LightGBM) Classification}
LightGBM is a gradient boosting framework designed for efficiency and scalability. It builds decision trees leaf-wise rather than level-wise, allowing for faster training and better accuracy, especially on large datasets like the credit default data. Light GBM model follows the standard boosting as in the ones given in (\ref{XG1}) and (\ref{h7}). LightGBM introduces histogram-based decision tree learning and gradient-based one-side sampling (GOSS) to reduce computational cost and memory usage. It also supports categorical features natively, enhancing performance without heavy preprocessing. Its speed and predictive power make it highly suitable for classification tasks involving credit default prediction.

\newpage
\subsection{Model Performance Metrics}	
All six (6) models were evaluated mainly using the test G-mean, F1 score, 
area under the receiver operating curve (ROC-AUC) and area under the precision-recall (PR-AUC), a type of precision-recall metric used in many classification settings, particularly where the datasets suffer class imbalance.\\

\noindent
The computational formulas for  specificity, recall, F1-score, ROC-AUC and PR-AUC are given respectively by (\ref{P1})-(\ref{PRAUCEQ}):

\begin{equation}
\text{Specificity}=\frac{\mathrm{TN}}{\mathrm{TN}+\mathrm{FP}}
	\label{P1}
\end{equation}

\begin{equation}
\text{Recall(Sensitivity)}=\frac{\mathrm{TP}}{\mathrm{TP}+\mathrm{FN}}
	\label{R1}
\end{equation}

\begin{equation}
	\text{F1-Score} = 	\frac{2(\mathrm{Precision} \times \mathrm{Recall})}{\mathrm{Precision} + \mathrm{Recall}}	
	\label{F1}
\end{equation}

\begin{equation}
\mathrm{ROC-AUC} = \int_{0}^{1} \mathrm{TPR}(\mathrm{FPR}) \, d(\mathrm{FPR})
	\label{AUCEQ}
\end{equation}

\begin{equation}
	\mathrm{PR-AUC} = \int_{0}^{1} \mathrm{Precision}(\mathrm{Recall}) \, d(\mathrm{Recall})
	\label{PRAUCEQ}
\end{equation}
where TP is the true positive rate, TN is the true negative rate, FP is the false positive rate and FN is the false negative rate.\\

\noindent
Likewise, the study also employed the G-mean statistic given in (\ref{G1}),
which is also an excellent indicator that effectively handles imbalanced class issues.
\begin{equation}
	\text{G-mean}=\sqrt{\text{(Recall/Sensitivity)} \times \text{Specificity}}	
	\label{G1}
\end{equation}
The G-mean is the geometric mean value used to measure overall model performance especially for imbalanced data as it is the case of this study. Poor classification results will produce a small G-mean value \cite{agyemang2025addressing}.\\

\begin{figure}[!hbtp]
	\centering
	\includegraphics[width=0.80\linewidth]{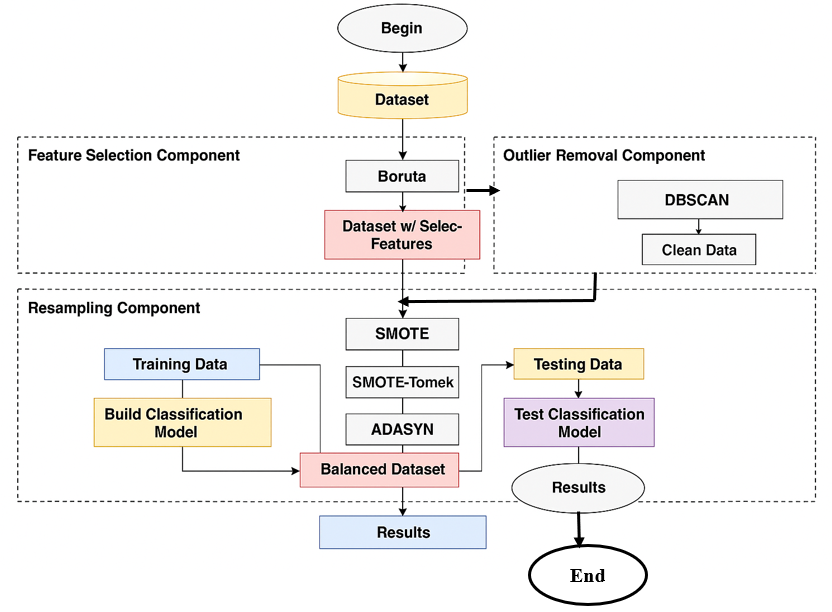}
	\caption{\textbf{General Architecture of the Model Building Process}}
	\label{Flowchart}
\end{figure}

\noindent
Figure \ref{Flowchart} depicts the general workflow employed for the study.

\newpage
\section{Results \& Discussion \label{Sec3}}
In this section, we present the results and findings of the simulation study.

\subsection{Descriptive statistics of quantitative variables}
\begin{figure}[!hbtp]
	\centering
	\includegraphics[width=0.70\linewidth]{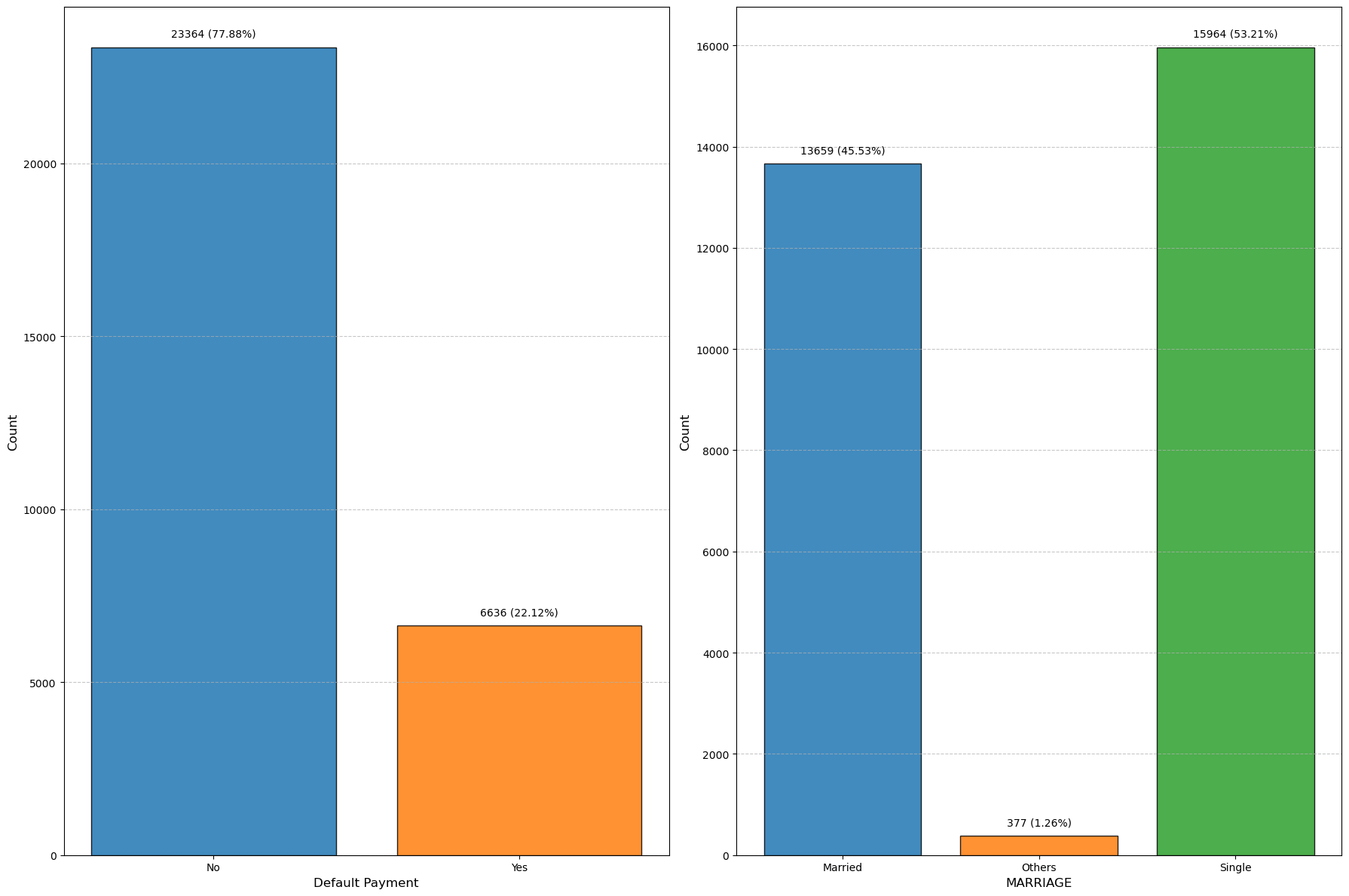}
	\caption{\textbf{Bar Charts showing the distribution of quantitative variables in the study}}
	\label{Bar1}
\end{figure}

\noindent
From Figure \ref{Bar1}, it is observed that majority of individuals - 23364 (77.88\%) did not default on credit payment, while 6636 (22.12\%) defaulted on credit payment. Regarding marital status, most individuals - 15964 (53.21\%) of participants are single, 13659 (45.53\%) are married, and 377 (1.26\%) identified as other form of marital status.\\

\color{black}
\noindent
The DBSCAN clustering algorithm was applied to the dataset after scaling all numeric variables using the StandardScaler. DBSCAN was run with an epsilon value of 0.5 as the neighborhood size and a minimum sample size of 5 indicating that 5 data points must be in a neighborhood to form a cluster. Each data point was assigned a cluster label, and outliers were identified as points with a label of -1. Outliers detected by DBSCAN were removed from the dataset prior to further analysis.
DBSCAN detected and removed 16480 outliers leaving 13520 samples with 10572 (78.20\%) non-default and 2948 (21.80\%) default. After DBSCAN was successfully applied, the class distribution of credit default still indicates a mild sign of class imbalance problem. First, the study fitted ML models for the Boruta+DBSCAN data without any resampling technique and compared their performances. Additionally, the study adopted three different oversampling techniques such as SMOTE, SMOTE-Tomek and ADASYN to balance the data for classification purposes. The study then assessed the predictive power of the machine learning algorithms considered in the study with Boruta selected features, DBSCAN outlier free samples coupled with  the three (3) oversampling techniques adopted.

\newpage
\subsection{Model Performance Evaluation of Data without Resampling}
Tables \ref{Hyper} consists of the model performance evaluation of all the six (6) machine learning models based on the training and testing data without the application of any resampling technique. For the purpose of this study, this model will be referred to as ``Baseline model". The training performance of the six (6) ML models were  assessed based on recall, specificity, F1-score and G-mean while the testing performance of the six (6) ML models were  assessed based on F1-score, G-mean, ROC-AUC and PR-AUC.\\

\begin{table}[htbp!]
	\centering
	\caption{\textbf{Model Comparison Results of Training and Testing Data for Boruta+DBSCAN}}
	\begin{tabular}{lcccccc}
		\toprule
		\textbf{Models} & \textbf{Recall} & \textbf{Specificity} & \textbf{F1-Score} & \textbf{G-mean} &\textbf{ROC-AUC} &\textbf{PR-AUC}\\ \hline
		\textbf{Training Phase}	&&&&&&\\ \hline
		Naive Bayes & 0.7884& 0.3773 & 0.3920&0.5454&&\\
		\vspace{0.02in}
		KNN & 0.9389& 0.9889 & 0.9490 &0.9636&&\\ \vspace{0.02in}
		Extreme Gradient Boosting & 0.3422& 0.9710 & 0.4733 &0.5765&&\\
		\vspace{0.02in}
		Adaptive Boosting & 0.2553& 0.9650 & 0.3698 &0.4964&&\\
		\vspace{0.02in}
		Gradient Boosting Machines (GBM) &  0.5246& 0.9853 & 0.6652 & 0.7190&&\\ 
		\vspace{0.02in}
		Light GBM & 0.3219& 0.9697 & 0.4500 &0.5587&&\\  \hline
		\textbf{Testing Phase}	&&&&&&\\ \hline
		Naive Bayes &0.7627& 0.3917 & 0.3869 &0.5466&0.6048&0.2907\\
		\vspace{0.02in}
		KNN & 0.2373& 0.8690 & 0.2781 &0.4541&0.5827&0.2948\\ \vspace{0.02in}
		Extreme Gradient Boosting & 0.2661& 0.9570 & 0.3747 &0.5046&0.7514&0.5024\\
		\vspace{0.02in}
		Adaptive Boosting & 0.2475& 0.9603 & 0.3561 &0.4875&0.7505&0.4854\\
		\vspace{0.02in}
		Gradient Boosting Machines (GBM) & 0.2831& 0.9461 & 0.3835 &0.5175&0.7370&0.4653\\ 
		\vspace{0.02in}
		Light GBM & 0.2695& 0.9579 & 0.3795 &0.5081&0.7488&0.5020\\
		\bottomrule
	\end{tabular}
	\label{Hyper}
\end{table}

\begin{figure}[!hbtp]
	\centering
	\includegraphics[width=0.95\linewidth]{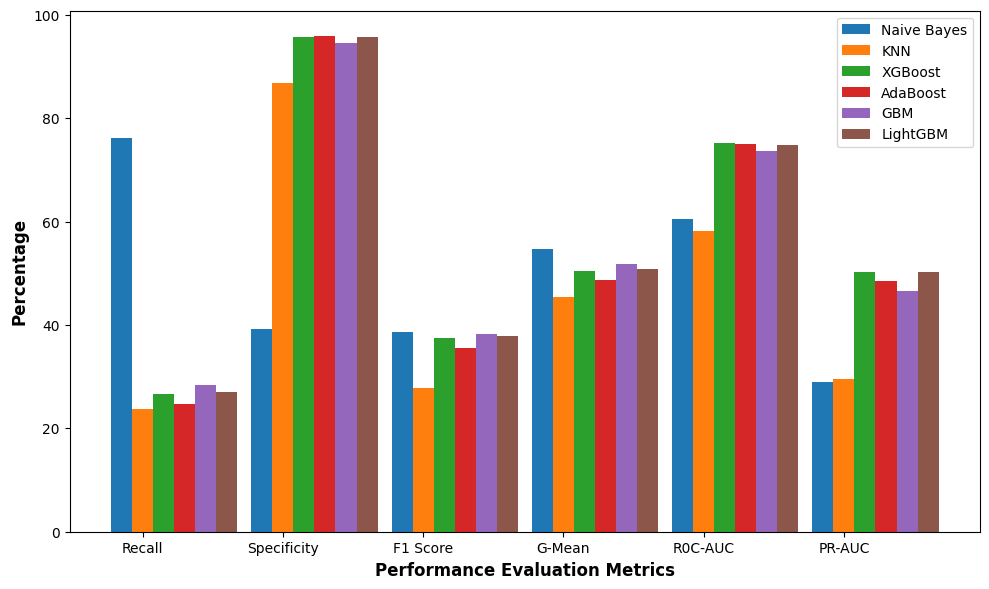}
	\caption{\textbf{Comparison of Baseline Models Based on Testing Performance Evaluation Metrics}}
	\label{BaselineModel}
\end{figure}

\noindent
From Table \ref{Hyp} and Figure \ref{BaselineModel}, it is evident that KNN outperforms the other models in the training phase (recall of 0.9389, specificity of 0.9889, F1-score of 0.9490 and a G-mean of 0.9636). However, in the testing phase, ensemble boosting algorithms such as XGBoost, AdaBoost, GBM and Light GBM performed better than KNN after the consideration of evaluation metric such as F1-score, G-mean, ROC-AUC and PR-AUC. 
The low testing metric values of the ML algorithms highlight that the mild class imbalance problem initially noted is adversely affecting model performance. To address this, the study employed resampling techniques such as SMOTE, SMOTE-Tomek and ADASYN to improve classification results.

\newpage
\color{black}
\subsection{Application of SMOTE, SMOTE-Tomek and ADASYN Oversampling Techniques}
\begin{figure}[!hbtp]
	\centering
	\subfigure[Class Distribution after SMOTE]
	{ \includegraphics[width=0.45\linewidth]{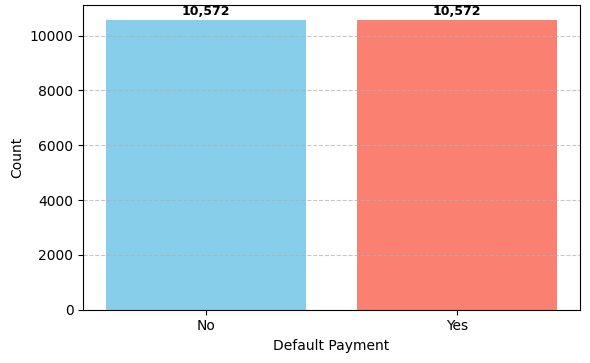}
		\label{D11}
	}
	\subfigure[Class Distribution after  SMOTE-Tomek]
	{ \includegraphics[width=0.45\linewidth]{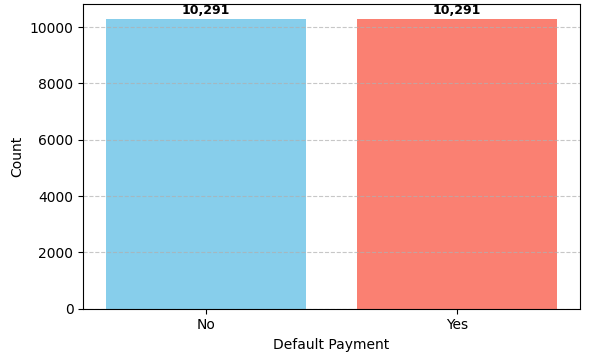}
		\label{D12}
	}
	\subfigure[Class Distribution after ADASYN]
	{ \includegraphics[width=0.45\linewidth]{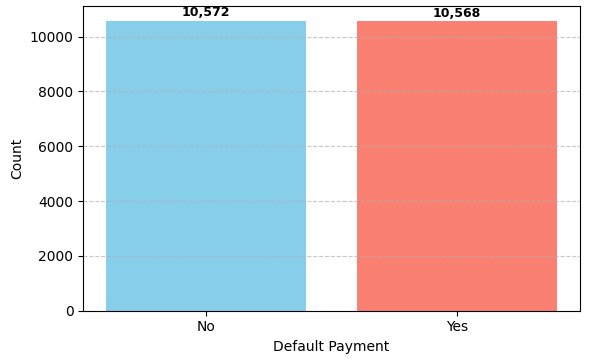}
		\label{D13}
	}
	\caption{\textbf{Bar Charts showing class distribution of Credit Default counts after Application of SMOTE, SMOTE-Tomek and ADASYN Oversampling Techniques}}
	\label{DDD}
\end{figure}

\noindent
From Figure \ref{DDD}, it was evident that undersampling techniques might not be ideal for the credit default data under consideration because it would involve removing a large number of instances from the majority class (non-default), which is significantly larger than the minority class (default ). This abnormal significant reduction may cause a loss of important information and produce models with poor generalization abilities. In lieu of the above disadvantage of applying an undersampling technique, the study rather adopted techniques that oversamples the minority class (default) by creating synthetic instances to balance the data without discarding any existing data.\\

\noindent
Figure \ref{DDD} presents the final samples after the application of the three (3) oversampling techniques: SMOTE, SMOTE-Tomek and ADASYN.  Figure \ref{D11} shows the final sample of SMOTE, which creates an equal number of 10572 cases for both default and non-default. Figure \ref{D12} presents the SMOTE combined with Tomek links oversampling technique, achieving a balanced dataset with 10291 instances in each class which is indicative that there were 281 instances that form Tomek links which were removed from SMOTE to generate the SMOTE-Tomek balanced sample. Tomek linkages were successfully used as a data cleaning strategy to get rid of samples produced by the SMOTE method that were close to the classification borderline. Figure \ref{D13} gives the ADASYN technique, which balances the credit default data by generating 10572 non-default and 10568 credit defaults instances, demonstrating a slight decrease in the number of synthetic samples for the minority class. 

\newpage
\subsection{Model Performance Evaluation of Resampled Data}
Tables \ref{Hyp}-\ref{Hyp2} consist of the model performance evaluation of all the six (6) machine learning models based on the training and testing data of the three (3) balancing techniques (SMOTE, SMOTE-Tomek and ADASYN respectively) adopted in the study. The training performance of the six (6) ML models were  assessed based on recall, specificity, F1-score and G-mean while the testing performance of the six (6) ML models were  assessed based on F1-score, G-mean, ROC-AUC and PR-AUC.\\

\noindent
For the SMOTE balanced Boruta+DBSCAN credit default data, in the training phase (Table \ref{Hyp}), KNN performed the best across all metrics with respective recall of 0.9836, specificity of 0.9742, F1-score of 0.9790 and a G-mean of 0.8938. GBM followed closely, showing strong performance (recall of 0.8897, specificity of 0.8968, F1-score of 0.8929 and a G-mean of 0.8932). Naive Bayes and AdaBoost had notably lower metrics in comparison, with Naive Bayes showing the least performance  with the lowest values across all metrics (specificity of 0.3333, F1-score of 0.6828 and a G-mean of 0.5366) except for recall of 0.8639. In Table \ref{Hyp}, following an initial training using the 5-fold cross-validation with F1-score as a scoring criteria coupled with extensive hyperparameter tuning, the testing phase produced interesting results. GBM outperformed the other five (5) ML models, with F1-score (0.8174), G-mean (0.8218), ROC-AUC (0.8973), and PR-AUC (0.9076), indicating better model discrimination ability for default and non-default. Light GBM also performed well in the testing phase, particularly in ROC-AUC (0.8837) and PR-AUC (0.8955). However, it is worth mentioning that Naive Bayes still lagged behind the other five (5) ML models considered in the testing phase. The statistical findings indicate that GBM outperformed KNN, Naive Bayes, XGBoost, AdaBoost and Light GBM in the testing phase when the Boruta+DBSCAN default data was balanced with SMOTE oversampling technique. 

\begin{table}[htbp!]
	\centering
	\caption{\textbf{Model Comparison Results of Training and Testing Data for Boruta+DBSCAN+SMOTE}}
	\begin{tabular}{lcccccc}
		\toprule
		\textbf{Models} & \textbf{Recall} & \textbf{Specificity} & \textbf{F1-Score} & \textbf{G-mean} &\textbf{ROC-AUC} &\textbf{PR-AUC}\\ \hline
		\textbf{Training Phase}	&&&&&&\\ \hline
	Naive Bayes & 0.8639& 0.3333 & 0.6828 &0.5366&&\\
		\vspace{0.02in}
		KNN & 0.9836& 0.9742 & 0.9790 & 0.9789 &&\\ \vspace{0.02in}
		Extreme Gradient Boosting & 0.8091& 0.8562 &0.8286 & 0.8323 &&\\ 
	\vspace{0.02in}
Adaptive Boosting & 0.7309& 0.8147 & 0.7629 & 0.7717 &&\\ 
\vspace{0.02in}
Gradient Boosting Machines (GBM) & 0.8897& 0.8968 & 0.8929 & 0.8932 &&\\ 
	\vspace{0.02in}
Light GBM & 0.8072& 0.8556& 0.8272 & 0.8310 &&\\  \hline
		\textbf{Testing Phase}	&&&&&&\\ \hline
		Naive Bayes &0.8538& 0.3329 & 0.6773 & 0.5331 &0.6618&0.6589\\
		\vspace{0.02in}
		KNN & 0.9120& 0.6652 & 0.8118 & 0.7789 &0.8579&0.8578\\ \vspace{0.02in}
		Extreme Gradient Boosting & 0.7673& 0.8326 & 0.7932 & 0.7993 &0.8795&0.8915\\ 
		\vspace{0.02in}
		Adaptive Boosting & 0.7275& 0.8123 & 0.7597 & 0.7687 &0.8476&0.8575\\ 
		\vspace{0.02in}
		Gradient Boosting Machines (GBM) &  0.7961& 0.8482 & 0.8174 & 0.8218 &0.8973&0.9076\\ 
		\vspace{0.02in}
		Light GBM & 0.7763& 0.8350 & 0.7997 & 0.8051 &0.8837&0.8955\\
		\bottomrule
	\end{tabular}
	\label{Hyp}
\end{table}

\noindent
For the SMOTE-Tomek balanced Boruta+DBSCAN default data, in the training phase (Table \ref{Hyp1}), KNN still remained the top performing model with the highest evaluation metrics with respective recall of 0.9721, specificity of 0.9789, F1-score of 0.9754 and a G-mean of 0.9755. GBM followed closely as it was the case for SMOTE balanced training data: with strong performance (recall of 0.9259, specificity of 0.9266, F1-score of 0.9262 and a G-mean of 0.9263). Naive Bayes and AdaBoost had notably lower metrics in comparison, with Naive Bayes showing the least performance  with the lowest values across all metrics except recall (specificity of 0.3337, F1-score of 0.6815 and a G-mean of 0.5361). Adaboost had the lowest recall of 0.6988 in the training phase.  Likewise, in Table \ref{Hyp1}, following an initial training using the 5-fold cross-validation with F1-score as a scoring criteria coupled with extensive hyperparameter tuning, GBM outperformed the other five (5) ML models, with F1-score (0.8256), G-mean (0.8298), ROC-AUC (0.9090), and PR-AUC (0.9185), indicating better model discrimination ability for default and non-default. Light GBM, a variant of GBM likewise performed well in the testing phase with an F1-score of 0.8068, G-mean of 0.8127, ROC-AUC (0.8912) and PR-AUC (0.9012). Here again, Naive Bayes still lagged behind the other five (5) ML models considered in the study with the least ROC-AUC of 0.6859 and PR-AUC of 0.6828. From a statistical point of view, the findings indicate that GBM outperformed KNN, Naive Bayes, XGBoost, AdaBoost and Light GBM in the testing phase when the Boruta+DBSCAN default data was balanced with SMOTE-Tomek oversampling technique as it was the case for Boruta+DBSCAN+SMOTE credit default data. 

\newpage
\begin{table}[htbp!]
	\centering
	\caption{\textbf{Model Comparison Results of Training and Testing Data for Boruta+DBSCAN+SMOTE-Tomek}}
	\begin{tabular}{lcccccc}
		\toprule
		\textbf{Models} & \textbf{Recall} & \textbf{Specificity} & \textbf{F1-Score} & \textbf{G-mean} &\textbf{ROC-AUC} &\textbf{PR-AUC}\\ \hline
		\textbf{Training Phase}	&&&&&&\\ \hline
		Naive Bayes & 0.8612& 0.3337& 0.6815 & 0.5361 &&\\
		\vspace{0.02in}
		KNN & 0.9721& 0.9789 & 0.9754 & 0.9755 &&\\ \vspace{0.02in}
		Extreme Gradient Boosting & 0.8109& 0.8612 & 0.8318 & 0.8356 &&\\ 
		\vspace{0.02in}
		Adaptive Boosting & 0.6988& 0.8133 & 0.7412 & 0.7539&&\\ 
		\vspace{0.02in}
		Gradient Boosting Machines (GBM) & 0.9259& 0.9266 & 0.9262 & 0.9263 &&\\ 
		\vspace{0.02in}
		Light GBM & 0.8171& 0.8590 & 0.8346& 0.8378 &&\\  \hline
		\textbf{Testing Phase}	&&&&&&\\ \hline
		Naive Bayes & 0.8717& 0.3545 & 0.6925 & 0.5559 &0.6859&0.6828\\
		\vspace{0.02in}
		KNN & 0.9009&0.6600 & 0.8040 & 0.7711 &0.8574&0.8627\\ \vspace{0.02in}
		Extreme Gradient Boosting & 0.7755& 0.8509 & 0.8059& 0.8123 &0.8911&0.9003\\ 
		\vspace{0.02in}
		Adaptive Boosting &  0.6992& 0.8213& 0.7446 & 0.7578 &0.8322&0.8369\\ 
		\vspace{0.02in}
		Gradient Boosting Machines (GBM) & 0.8042& 0.8562 & 0.8256 & 0.8298 &0.9090&0.9185\\ 
		\vspace{0.02in}
		Light GBM &0.7794& 0.8475 & 0.8068 & 0.8127 &0.8912&0.9012\\
		\bottomrule
	\end{tabular}
	\label{Hyp1}
\end{table}

\noindent
With the consideration of ADASYN balanced Boruta+DBSCAN credit default data, in the training phase (Table \ref{Hyp2}), KNN once again emerged as the top performing model with the highest evaluation metrics with respective recall of 0.9824, specificity of 0.9752, F1-score of 0.9788 and a G-mean of 0.9788. GBM followed closely as it was the case for SMOTE and SMOTE-Tomek balanced training data: with strong performance (recall of 0.9252, specificity of 0.9266, F1-score of 0.9258 and a G-mean of 0.9259). Naive Bayes and AdaBoost had notably lower metrics in comparison with the other four (4) ML models, with Naive Bayes showing the least performance  with the lowest values across all metrics except recall (specificity of 0.3177, F1-score of 0.6740 and a G-mean of 0.5212). Adaboost still had the least recall of 0.6859 in the ADASYN training phase.  Likewise, considering the testing phase in Table \ref{Hyp2}, GBM emerged as the top performer comparative to the other five (5) ML models, with F1-score (0.8148), G-mean (0.8185), ROC-AUC (0.8972), and PR-AUC (0.9082), indicative of better model discrimination ability for credit default and non-default. Naive Bayes still remained as the worst performer lagging behind the other five (5) ML models with the least F1-score of 0.6738, G-mean of 0.5240, ROC-AUC of 0.6514 and PR-AUC of 0.6531. It was concluded that GBM outperformed KNN, Naive Bayes, XGBoost, AdaBoost and Light GBM in the testing phase when the Boruta+DBSCAN default data was balanced with ADASYN oversampling technique as it was with Boruta+DBSCAN+SMTOTE and Boruta+DBSCAN+SMOTE-Tomek credit default data. 

\begin{table}[htbp!]
	\centering
	\caption{\textbf{Model Comparison Results of Training and Testing Data for Boruta+DBSCAN+ADASYN}}
	\begin{tabular}{lcccccc}
		\toprule
		\textbf{Models} & \textbf{Recall} & \textbf{Specificity} & \textbf{F1-Score} & \textbf{G-mean} &\textbf{ROC-AUC} &\textbf{PR-AUC}\\ \hline
		\textbf{Training Phase}	&&&&&&\\ \hline
		Naive Bayes & 0.8552& 0.3177 & 0.6740 & 0.5212 &&\\
		\vspace{0.02in}
		KNN &  0.9824& 0.9752&0.9788 & 0.9788 &&\\ \vspace{0.02in}
		Extreme Gradient Boosting & 0.7948& 0.8542 & 0.8191 & 0.8240 &&\\ 
		\vspace{0.02in}
		Adaptive Boosting & 0.6859& 0.8133 & 0.7326 & 0.7469 &&\\ 
		\vspace{0.02in}
		Gradient Boosting Machines (GBM) & 0.9252& 0.9266 & 0.9258 & 0.9259 &&\\ 
		\vspace{0.02in}
		Light GBM & 0.7897& 0.8567 & 0.8170 & 0.8225 &&\\  \hline
		\textbf{Testing Phase}	&&&&&&\\ \hline
		Naive Bayes & 0.8524& 0.3221& 0.6738 & 0.5240&0.6514&0.6531\\
		\vspace{0.02in}
		KNN & 0.9210& 0.6339 & 0.8054 & 0.7641 &0.8483&0.8497\\ \vspace{0.02in}
		Extreme Gradient Boosting & 0.7512& 0.8321 & 0.7828& 0.7906 &0.8751&0.8850\\ 
		\vspace{0.02in}
		Adaptive Boosting & 0.6774& 0.8155 & 0.7276 & 0.7433&0.8132&0.8234\\ 
		\vspace{0.02in}
		Gradient Boosting Machines (GBM) & 0.7971& 0.8406 & 0.8148&  0.8185 &0.8972&0.9082\\ 
		\vspace{0.02in}
		Light GBM &0.7498& 0.8401 & 0.7852 & 0.7937 &0.8775&0.8883\\
		\bottomrule
	\end{tabular}
	\label{Hyp2}
\end{table}

\newpage
\color{black}
\subsection{Model Performance of GBM Model with SMOTE, SMOTE-Tomek and ADASYN}
\begin{figure}[!hbtp]
	\centering
	\includegraphics[width=0.85\linewidth]{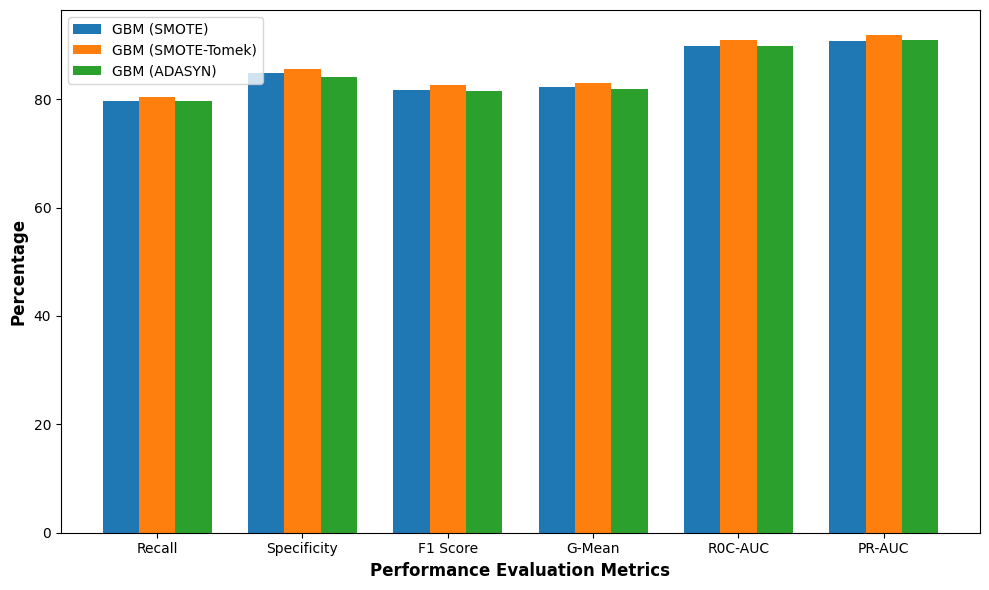}
	\caption{\textbf{Comparison of GBM Models Based on Testing Performance Evaluation Metrics}}
	\label{ModComp}
\end{figure}

\noindent
After rigorous assessment of all the six (6) machine learning models considered in this study, it was observed that the GBM model under SMOTE, SMOTE-Tomek and ADASYN were adjudged as the three (3) best models for classifying defaulters and non-defaulters for the credit default data. Figure \ref{ModComp} presents the graphical comparison of the three (3) best models under the three (3) different oversampling techniques using the testing data. We found Boruta+DBSCAN+SMOTE-Tomek+GBM (with ROC-AUC of 90.90\%, PR-AUC of 91.85\%, F1-score of 82.56\% and a G-mean of 82.98\%) to be superior compared to the other two models: Boruta+DBSCAN+SMOTE+GBM (with ROC-AUC of 89.73\%, PR-AUC of 90.76\%, F1-score of 81.74\% and a G-mean of 82.18\%) and Boruta+DBSCAN+ADASYN+GBM (with ROC-AUC of 89.72\%, PR-AUC of 90.82\%, F1-score of 81.48\% and a G-mean of 81.85\%) after the consideration of  imbalance performance evaluation metrics such as F1 score, G-mean, ROC-AUC and PR-AUC values.

\noindent 
The time to train each GBM model in seconds of the three (3) oversampling techniques was further assessed.

\begin{figure}[!hbtp]
	\centering
	\includegraphics[width=0.55\linewidth]{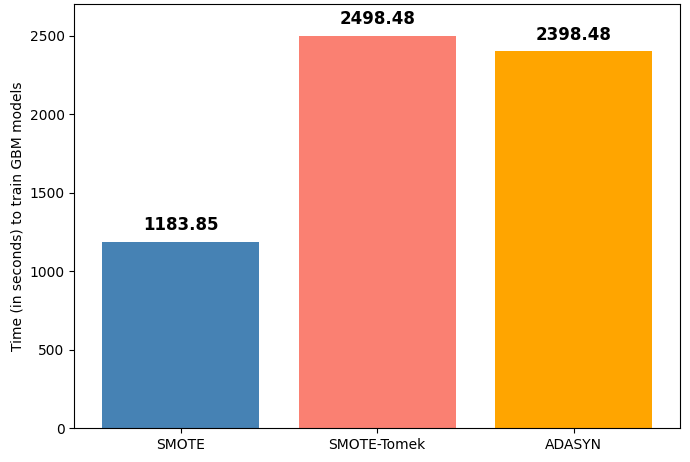}
	\caption{\textbf{Time to Train GBM Models with SMOTE, SMOTE-Tomek, and ADASYN}}
	\label{TrainTime}
\end{figure}

\noindent
Figure \ref{TrainTime} displays the time taken to train GBM models after applying three different resampling techniques: SMOTE, SMOTE-Tomek and ADASYN. SMOTE recorded the lowest training time at 1183.85 seconds. In contrast, SMOTE-Tomek and ADASYN required significantly longer times, at 2498.48 and 2398.48 seconds respectively. The increased times for SMOTE-Tomek and ADASYN suggest higher computational costs, likely due to their more complex resampling processes.

\newpage
\subsection{Gradient Boosting Model Hyperparameter Optimization}
To identify optimal hyperparameters, grid search with 5-fold cross-validation was performed for each GBM model using different resampling techniques on the default data. Table \ref{Tab4} presents the best-performing configurations after extensive hyperparameter tuning.

\begin{table}[H]
	\centering
	\caption{Best hyperparameters for GBM models with different resampling methods}
	\label{Tab4}
	\begin{tabular}{lcccccc}
		\toprule
		\textbf{Resampling Method} & \textbf{Fits} & \textbf{learning\_rate} & \textbf{max\_depth} & \textbf{max\_features} & \textbf{n\_estimators} & \textbf{subsample} \\
		\midrule
		SMOTE           & 360 & 0.1 & 7 & \texttt{sqrt} & 100 & 1.0 \\
		SMOTE+Tomek     & 540 & 0.1 & 7 & \texttt{None} & 150 & 0.5 \\
		ADASYN          & 540 & 0.1 & 7 & \texttt{None} & 150 & 1.0 \\
		\bottomrule
	\end{tabular}
\end{table}
\noindent
The results indicate that although the learning rate and tree depth were consistent across all models, other parameters such as \texttt{max\_features}, \texttt{n\_estimators}, and \texttt{subsample} varied depending on the resampling technique used. This suggests that resampling strategy influences not only model performance but also the optimal configuration of the GBM.

\subsection{ROC-AUC and PR-AUC of GBM with SMOTE, SMOTE-Tomek and ADASYN}
\begin{figure}[!hbtp]
	\centering
	\subfigure[ROC-AUC of GBM]
	{ \includegraphics[width=0.485\linewidth]{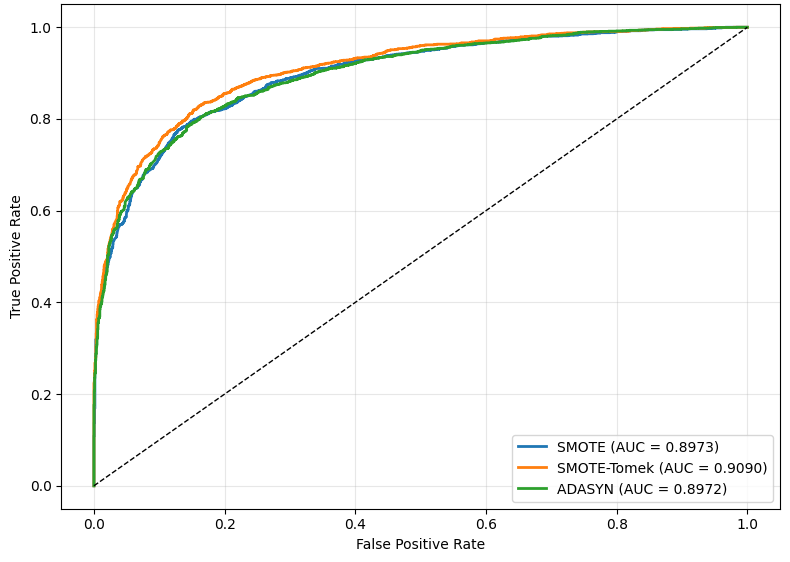}
		\label{AUC}
	}\hfill
	\subfigure[PR-AUC of GBM]
	{ \includegraphics[width=0.485\linewidth]{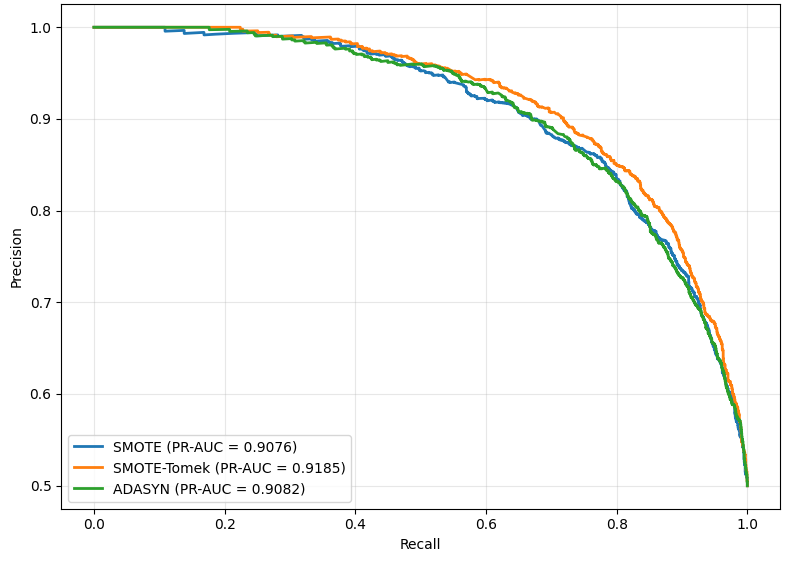}
		\label{PRAUC}
	}
	\caption{\textbf{AUC-ROC and AUC-PR for GBM of SMOTE, SMOTE-Tomek and ADASYN}}
	\label{AUCPRAUC}
\end{figure}

\noindent
Figures \ref{AUCPRAUC} depicts the discrimination performance of the GBM classifier under the three different class balancing techniques considered in this study: SMOTE, SMOTE-Tomek, and ADASYN. The evaluation was performed using both the ROC-AUC and the PR-AUC. As shown in Figure \ref{AUC}, the ROC  for all three oversampling strategies demonstrate strong model performance, with ROC-AUC values of 0.8973 for SMOTE, 0.9090 for SMOTE-Tomek, and 0.8972 for ADASYN. SMOTE-Tomek yielded the highest ROC-AUC, indicating a marginally better overall ability to distinguish between defaulters and non-defaulters. Figure \ref{PRAUC} presents the corresponding precision-recall curves. The PR-AUC was highest for SMOTE-Tomek (0.9185), followed by ADASYN (0.9082) and SMOTE (0.9076). The elevated PR-AUC values further highlight the effectiveness of the models in identifying true positive cases, particularly in the presence of class imbalance. From a statistical perspective, the Boruta+DBSCAN+SMOTE-Tomek provided a slight improvement in both ROC-AUC and PR-AUC compared to Boruta+DBSCAN+SMOTE and Boruta+DBSCAN+ADASYN, suggesting that it may be the most effective resampling technique for this credit default prediction task.\\

\noindent
As observed, the performance evaluation metrics of the six (6) ML models
are higher for our proposed models using over-sampling
techniques for the credit default dataset. The empirical results of this study show that the proposed models using imbalanced techniques (SMOTE, SMOTE-Tomek and ADASYN) with the Boruta DBSCAN outlier free data has significantly improved the performance from the baseline model (with improved testing F1-score, G-mean, ROC-AUC and PR-AUC). Our experimental results demonstrate that incorporating class balancing strategies into classification models for mild imbalance data enhances their robustness and reliability by improving the models' learning capacity for the minority class.  The results from the credit default dataset provide a significant enhancement in the testing F1-score, G-mean, ROC-AUC, and PR-AUC values.

\newpage
\color{black}
\section{Conclusion \& Recommendation \label{Sec4}}
This study evaluates the performance of several machine learning models, including Naive Bayes, KNN, XGBoost, AdaBoost, GBM and Light GBM, in predicting credit card default for financial service applications. A comprehensive examination of credit default prediction by comparing three techniques namely SMOTE, SMOTE-Tomek, and ADASYN that are commonly used to address the class imbalance problem in  credit default situations was conducted. Likewise, recognizing that credit default datasets are typically skewed, with defaulters comprising a much smaller proportion than non-defaulters, we began our analysis by evaluating machine learning (ML) models on the imbalanced data without any resampling to establish baseline performance. These baseline results provide a reference point for understanding the impact of subsequent balancing methods. Among the models tested, Boruta+DBSCAN+SMOTE-Tomek+GBM (with ROC-AUC of 90.90\%, PR-AUC of 91.85\%, F1-score of 82.56\% and a G-mean of 82.98\%) demonstrated the best performance compared to the other two best models: Boruta+DBSCAN+SMOTE+GBM (with ROC-AUC of 89.73\%, PR-AUC of 90.76\%, F1-score of 81.74\% and a G-mean of 82.18\%) and Boruta+DBSCAN+ADASYN+GBM (with ROC-AUC of 89.72\%, PR-AUC of 90.82\%, F1-score of 81.48\% and a G-mean of 81.85\%) after the consideration of  imbalance performance evaluation metrics such as F1 score, G-mean, ROC-AUC and PR-AUC values. This indicates that GBM has the highest ability to accurately differentiate between credit defaulters and non-credit defaulters, making it an excellent choice for credit default management in financial systems.\\

\noindent
However, a shocking revelation was the fact that even though KNN performed excellently during the training phase: Boruta+DBSCAN+SMOTE+KNN (with recall of 0.9836, specificity of 0.9742, F1-score of 0.9790 and a G-mean of 0.8938), Boruta+DBSCAN+SMOTE-Tomek+KNN (with recall of 0.9721, specificity of 0.9789, F1-score of 0.9754 and a G-mean of 0.8356) and Boruta+DBSCAN+ADASYN+KNN (with recall of 0.9824, specificity of 0.9752, F1-score of 0.9788 and a G-mean of 0.9788) in all the three (3) oversampling schemes, it failed to perform well in the testing phase comparative to GBM. In credit default applications, characterized by big data and the necessity for real-time decision-making, ensemble boosting algorithms such as Boruta+DBSCAN+SMOTE-Tomek+GBM is distinguished in this study as the most efficient model owing to its high performance evaluation metrics in the testing phase, and capacity to manage intricate, imbalanced datasets.\\

\noindent
The application of ML models to predict imbalance credit card default and or credit scoring data are well documented in the literature. For instance, \cite{brown2012experimental} compared several classification algorithms on real-world credit scoring data sets with different levels of class imbalance. It was observed that Random forests and gradient boosting consistently outperformed other models, especially when the number of defaulters was very small compared to non-defaulters. These ensemble methods maintained high predictive accuracy even in extremely imbalanced scenarios. In contrast, classic algorithms like decision trees (C4.5), quadratic discriminant analysis (QDA), and k-nearest neighbors struggled as the data became more imbalanced. The study highlights the particular strength of  boosting algorithms for imbalanced credit scoring problems. Likewise, the study by \cite{hossain2025comparative} compared the performance of five machine learning models including Logistic Regression, Random Forest, XGBoost, Support Vector Machines, and Neural Networks for predicting credit risk in banking systems. Among the models, XGBoost demonstrated the highest accuracy (88.70\%), precision (89.50\%), recall (80.30\%), and AUC (91.30\%), making it the most effective at identifying high-risk borrowers. Random Forest and Neural Networks also performed well but were less optimal in either interpretability or computational cost. The boosting algorithm,  XGBoost was recommended for large-scale applications due to its predictive strength and ability to handle complex, imbalanced data. The empirical results of \cite{brown2012experimental} and \cite{hossain2025comparative} are in direct agreement with the findings of this study (as evident in Table \ref{Hyp}, Table \ref{Hyp1} and Tables \ref{Hyp2}).\\

\noindent
 In this study, several contributions to the field of credit default prediction are made. First, the combination of Boruta feature selection and DBSCAN algorithms provide practitioners and researchers with credit default data free of irrelevant features and outliers for robust prediction. Second,  a detailed comparison of the performance of various oversampling techniques including SMOTE, SMOTE-Tomek and ADASYN on the performance of machine learning models are assessed to identify which combinations of models and oversampling techniques yield the best outcomes for predicting credit default. This comprehensive evaluation (encompassing the application of Boruta, DBSCAN, SMOTE, SMOTE-Tomek, ADASYN and several ML models) offers valuable insights for improving credit default prediction in the long run.\\
 
 \noindent
 Although the three GBM models (Boruta+DBSCAN coupled with SMOTE, SMOTE-Tomek and ADASYN) showed strong predictive performance in predicting credit defaults, their longer training times, especially with more complex resampling methods like SMOTE-Tomek (2498.48 seconds) and ADASYN (2398.48 seconds), have practical implications. Extended model runtimes can lead to increased computational costs and may limit how quickly results can be delivered, which is a concern in real-time or resource-constrained environments. Therefore, while GBM models are highly effective for predicting credit default, it is important for practitioners to balance model accuracy with efficiency, taking into account the available computational resources and the urgency of decision-making when deploying these models in operational settings. Future study could concentrate on enhancing the balance between accuracy and interpretability in credit default modeling, alongside investigating the incorporation of these models into real-time banking systems to facilitate the credit approval process and augment credit risk management.

\subsection*{Declaration of competing interest}
\noindent
The authors declare that they have no known competing financial interests or personal relationships that could have appeared to influence the work reported in this paper.

\subsection*{Author Contributions}
All authors declare to have contributed equally to this manuscript. All authors read and approved 
of the final manuscript submitted for publication. 

\subsection*{Data Availability}
\noindent
The data used to support the findings of this study are available upon reasonable request from the corresponding author.

\subsection*{Acknowledgment}
\noindent
The second and corresponding author  acknowledges the enormous support of the University of Tulane Dean Research  Council Scholarship and the University of Texas Rio Grande Valley (UTRGV) Presidential Research Fellowship Fund.

\bibliographystyle{unsrt} 

\bibliography{REFERENCES} 
\end{document}